\documentclass[runningheads,a4paper]{llncs}

\usepackage{amssymb}
\usepackage{float}
\usepackage[noend]{algpseudocode}
\usepackage[english]{my-shortcuts}
\usepackage{graphicx}   % Package de l'AMS

\usepackage{url}

\newcommand{\keywords}[1]{\par\addvspace\baselineskip
	\noindent\keywordname\enspace\ignorespaces#1}

\begin{document}
	
	\mainmatter  % start of an individual contribution
	
	\title{Feature selection in weakly coherent matrices}
	
	\titlerunning{Incoherent perturbation}
	
	\author{St\'ephane Chr\'etien \and Olivier Ho}
	\institute{National Physical Laboratory,\\
		Hampton Road, TW11 0LW Teddington, UK\\
		\url{https://sites.google.com/view/stephanechretien/home}
		\\
		and \\ 
		Universit\'e de Franche Comt\'e,\\
		16 route de Gray, 25000 Besan{\c c}on, France \\}

	
	\toctitle{Lecture Notes in Computer Science}
	\tocauthor{Authors' Instructions}
	\maketitle

	\begin{abstract}
		A problem of paramount importance in both pure (Restricted Invertibility problem) and applied mathematics (Feature extraction) is the one of selecting a submatrix of a given matrix, such that this submatrix has its smallest singular value above a specified level. Such problems can be addressed using perturbation analysis. In this paper, we propose a  perturbation bound for the smallest singular value of a given matrix after appending a column, under the assumption that its initial coherence is not large, and we use this bound to derive a fast algorithm for feature extraction.

	\end{abstract}

	\section{Introduction}
	{\em In this paper, all considered matrices will be assumed to have their columns $\ell_2$-normalised.}
	
	\subsection{Background on singular value perturbation}
	
	Spectrum perturbation after appending a column has been addressed recently in the literature as a key ingredient in the study of graph sparsification \cite{batson2013spectral}, control of pinned systems of ODE's \cite{porfiri2008criteria}, the spiked model in statistics \cite{nadler2008finite}; it can also be useful in Compressed Sensing \cite{chretien2014perturbation} or for the column selection problem \cite{chretien2015elementary}. It is also connected to column selection problems in pure mathematics (Grothendieck and Pietsch factorisation and the Bourgain-Tzafriri restricted invertibility problem) \cite{tropp2009column}. 
	
	The goal of the present paper is to study this particular perturbation problem in the special context of column subset selection. The column selection problem was proved essential in High Dimensional Data Analysis \cite{mahoney2009cur}, \cite{zhao2007spectral}, 
	\cite{ben2003detecting}, \cite{krzanowski1987selection}. \cite{wolf2005feature}, etc. Different criteria for column subsect selection have been studied \cite{boutsidis2014near}. The need for efficient column selection in the era of Big Data is more pressing than ever. Moreover, deterministic techniques are often prefered over randomised techniques in industrial applications due to repetability constraints. 
	
	\subsection{Previous approaches to column selection}
	
	Several approaches have been extensively discussed in the literature. Other \textit{deterministic} approaches have beed studied recently in the pure mathematics literature, namely \cite{spielman2012elementary}, \cite{youssef2014restricted}. However, these approaches are computationally expensive because of the necessity to perform a matrix inversion at each step. The method of \cite{tropp2009column} combines randomness with semi-definite programming and although very elegant, is not computationally efficient in practice. A quite efficient techniques is the rank-revealing QR decomposition. Table 1 in \cite{boutsidis2009improved} provides the performance of this approach and compares it with various other methods. Randomized sampling-based approaches sometimes prove to be faster than the deterministic approaches. For instance methods based e.g. on leverage scores is often giving satisfactory results in practice. Note also that CUR decomposition is much related to the  Column Selection tasks and the associated methods can be relevant in practice. A very interesting and efficient approach is the simple greedy algorithm presented in \cite{farahat2011efficient} and \cite{farahat2011efficient}. However, the method of 
	\cite{farahat2013efficient} does not allow for control on the smallest singular value of the selected submmatrix, a criterion which often considered important for selecting sufficiently decorrelated features.

	\subsection{Coherence}
	The coherence of a matrix $X$, usually denoted by $\mu(X)$, is defined as
	\bea
	\mu (X)& = & \max_{1\le k<l \le p} |\la X_{k},X_{l} \ra|.
	\eea
	If the coherence is equal to zero, then the matrix is orthogonal. On the other hand, small coherence does not mean that $X$  is close to square and orthogonal. Indeed, as easy computations show, e.g. i.i.d. Gaussian matrices with values in $\mathbb R^{n\times p}$ and normalised columns can have a coherence of order $\log(p)^{-1}$ even for $n$ of order $\log(p)^3$; see \cite[Section 1.1]{candes2009near}. Situations where small coherence holds arise often in practice, especially in signal processing \cite{candes2007sparsity} and statistics \cite{candes2009near}. 
	The coherence of a matrix has attracted renewed interest recently due to its prominent role in Compressed Sensing \cite{candes2008introduction}, Matrix Completion \cite{recht2011simpler}, Robust PCA \cite{candes2011robust} and Sparse Estimation in general. The relationship between coherence and how many columns one can extract uniformly at random which build up a robustly invertible submatrix are studied in \cite{chretien2012invertibility}. When the coherence is not sufficiently small, the results in \cite{chretien2012invertibility} are not so much useful anymore and we should turn to the problem of extracting one submatrix with largest possible number of columns with smallest possible correlation. Using coherence information in the study of fast column selection procedures is one interesting question to address in this field.
	
	\subsection{Contribution of the paper}
	We propose a greedy algorithm for column subset selection and apply this algorithm to some practical problems.
	Our contribution to the perturbation and the column selection problems focuses on the special setting where the matrix under study has low coherence. Interestingly, standard perturbation results, e.g. \cite{bhatia2007perturbation} do not take into account the potential incoherence of the matrix under study. The results presented in this paper seem to be the first to incoporate such prior information into the analysis of a column subset selection procedure. 
	
	Our approach here is based on a new eigenvalue perturbation bound for matrices with small coherence. 
	Previous bounds have been obtained using the famous Gershgorin's circles theorem 
	\cite{bandeira2013road} but Gershgorin's bound is often to crude as demonstrated in \cite{chretienhoprep}. and recent advances have been obtained in this direction in \cite{spielman2012elementary} and \cite{youssef2014restricted}.

	\section{Main results}
	\label{main}
	
	Our main result is a bound on the smallest singular value after appending a column of a given data matrix with potentially small coherence. Our approach is based on a new result about eigenvalue perturbation. Perturbation after appending a column is a special 
	type of perturbation \cite{chretien2014perturbation}. The goal of the next subsections is to prove refined results of this type for this problem. 
	
	Theorem \ref{theorem:2} is our first main result on perturbation. This result gives a perturbation bound on the spectrum of a submatrix $X_{T_0}$ of a matrix $X$.
	Corollary \ref{cor3.3} takes into account the fact that the coherence of a submatrix can be smaller by a factor $\alpha$ than the coherence of the full matrix. This factor $\alpha$ is crucial in the study of e.g. greedy algorithms for column selection where at each step, the selected submatrix has much better coherence than the full matrix from which it is extracted. Corollary \ref{cor3.4} proves a bound on the smallest singular value after successively appending several columns. An example where this result will be usefull is the application to greedy column selection algorithms where it can provide a relevant stopping criterion.

	\subsection{Appending one vector: perturbation of the smallest non zero eigenvalue}
	If we consider a subset $T_0$ of $\{1,\ldots,p\}$ and a submatrix $X_{T_0}$ of $X$, the problem of studying the eigenvalue perturbations resulting from appending a column $X_j$ to $X_{T_0}$, with $j\not\in T_0$ 
	can be studied using Cauchy's Interlacing Lemma as in the following result. 
	\begin{theo} \label{theorem:2}
		Let $T_0\subset \{1,\ldots,p\}$ with $\vert T_0 \vert=s_0$ and $X_{T_0}$ a submatrix of $X$. Let  $\lambda_1\left(X_{T_0}X_{T_0}^t\right)\geq ... \geq \lambda_{s_0}\left(X_{T_0}X_{T_0}^t\right)$ be the eigenvalues of $X_{T_0}X^t_{T_0}$. 
		%Assume that %$0<\min_{i=1}^{s_0} \lambda_i(X_{T_0}^tX_{T_0}) <1$ and that
		%$\lambda_{s_0}< 1-s_0\mu^2$. 
		We have
		\begin{align}
		\lambda_{s_0+1} \left(X_{T_0}X_{T_0}^t+X_jX_j^t\right) & \ge   \lambda_{s_0}\left(X_{T_0}X_{T_0}^t\right)- \min \left(\Vert X_{T_0}^t X_j\Vert_2, \frac{\Vert X_{T_0}^t X_j \Vert^2_2}{1-\lambda_{s_0}\left(X_{T_0}X_{T_0}^t\right)}\right). 
		\label{lema2}
		\end{align}
		
	\end{theo} 	
	\begin{proof}
		Setting $v=X_j$
		\begin{align*}
		A & = X_{T_0} X_{T_0}^t 
		\end{align*}
		we obtain from Proposition \ref{interlace} that the smallest nonzero eigenvalue of $X_{T_0}X_{T_0}^t+X_jX_j^t$ is the smallest root of 
		\begin{align*}
		f(x) & = 1-\sum_{i=1}^n \frac{\langle v,u_i \rangle^2}{x - \lambda_i\left(X_{T_0}X_{T_0}^t\right)}.
		\end{align*}
		We can decompose this function into two terms
		\begin{align*}
		f(x) & = 1-\sum_{i=1}^{s_0} \frac{\langle v,u_i \rangle^2}{x - \lambda_i\left(X_{T_0}X_{T_0}^t\right)}- 	\sum_{i=s_0+1}^{n} \frac{\langle v,u_i \rangle^2}{x - \lambda_i\left(X_{T_0}X_{T_0}^t\right)}.
		\end{align*}
		Since $\lambda_i\left(X_{T_0}X_{T_0}^t\right)=0$ for $i=s_0+1,\ldots,n$, we get 
		\begin{align*}
		f(x) & = 1+\sum_{i=1}^{s_0} \frac{\langle v,u_i \rangle^2}{ \lambda_i\left(X_{T_0}X_{T_0}^t\right)-x}- 	\sum_{i=s_0+1}^{n} \frac{ \langle v,u_i \rangle^2}{x}.
		\end{align*}
		Notice that
		\begin{align*}
		\sum_{i=1}^{s_0} \langle v,u_i\rangle^2 & \le \frac1{\lambda_{s_0}\left(X_{T_0}X_{T_0}^t\right)} 
		\sum_{i=1}^{s_0} \lambda_{i}\left(X_{T_0}X_{T_0}^t\right) \ \langle v,u_i\rangle^2 = \frac1{\lambda_{s_0}\left(X_{T_0}X_{T_0}^t\right)} \Vert X_{T_0}^tv\Vert_2^2. 
		\end{align*}
		Since $f$ is increasing on the set $]0,\lambda_{s_0}\left(X_{T_0}X_{T_0}^t\right)[$,
		the smallest root of $f$ is larger than the smallest positive root of $\tilde f$ with
		\begin{align*}
		\tilde f(x) & = 1+\
		\frac{\Vert X_{T_0}^t X_j \Vert^2_2}{ \lambda_{s_0}\left(X_{T_0}X_{T_0}^t\right) (\lambda_{s_0}\left(X_{T_0}X_{T_0}^t\right)-x)}
		-\frac{1-\lambda_{s_0}\left(X_{T_0}X_{T_0}^t\right)^{-1}\Vert X_{T_0}^t X_j \Vert^2_2}{x}.
		\end{align*}
		Thus, after some easy calculations,
		we find that 
		\begin{align*}
		\lambda_{s_0+1} \left(X_{T_0}X_{T_0}^t+X_jX_j^t\right) & \ge 
		\frac{1+\lambda_{s_0}\left(X_{T_0}X_{T_0}^t\right)-\sqrt{(1-\lambda_{s_0}\left(X_{T_0}X_{T_0}^t\right))^2+4 \Vert X_{T_0}^t X_j \Vert^2_2}}2 
		\end{align*}
		which, using  $\sqrt{a+b}\le \sqrt{a}+\sqrt{b}$ and $\sqrt{1+a}\le 1+\frac{a}{2}$,
		easily gives \eqref{lema2}.

	\end{proof}  
	
	This theorem is useful in the case where $\mu$ small enough so that $\Vert X_{T_0}^t X_j \Vert^2_2 \leq 1$.
	In practice, the submatrices $X_{T_0}$ of $X$ have better coherence than $X$, up to a factor $\alpha$. Moreover, we have $\Vert X_{T_0}X_j\Vert_2^2 \le s_0 \mu^2$. The following corollary rephrases Theorem \ref{main} using the parameter $\alpha$.
	\begin{cor}\label{cor3.3}
		Let $X$ and $T_0$ be defined as in Theorem \ref{theorem:2} and assume
		\[\Vert X_{T_0}^t X_j\Vert_2^2 \le \alpha s_0 \mu^2. \]
		Then
		\begin{align}
		\lambda_{s_0+1} \left(X_{T_0}X_{T_0}^t+X_jX_j^t\right) & \ge   \lambda_{s_0}\left(X_{T_0}X_{T_0}^t\right)- \min \left(\sqrt{\alpha s_0 \mu^2}, \frac{\alpha s_0 \mu^2}{1-\lambda_{s_0}\left(X_{T_0}X_{T_0}^t\right)}\right). 
		\end{align}
	\end{cor}
	\subsection{Successive perturbations}
	If we append $s_1$ columns successively to the matrix $X_{T_0}$, we obtain the following result
	\begin{cor}\label{cor3.4}
		Let $T_0\subset \{1,\ldots,p\}$ with $\vert T_0 \vert=s_0$ and $X_{T_0}$ a submatrix of $X$. Let $T_1\subset \{1,\ldots,p\}$ with $\vert T_1 \vert=s_1$ and $T_0 \cap T_1=\emptyset$. Let
		\begin{align}
		\label{epsmin}
		\varepsilon_{min} & =\min\left(\sqrt{\alpha \mu^2} \sum\limits_{i=s_0}^{s_0+s_1} \sqrt{i},\frac{\alpha \mu^2 s_0}{1-\lambda_{s_0}\left(X_{T_0}X_{T_0}^t\right)}+\frac{2(1-\lambda_{s_0}\left(X_{T_0}X_{T_0}^t\right))}{s_0}\sum_{i={s_0}+1}^{ s_0+s_1}\frac{i}{i-1}\right).
		\end{align}
		Then
		\begin{align}
		\lambda_{s_0+s_1} \left(X_{T_0 \cup T_1}^t X_{T_0 \cup T_1}  \right) & \ge \lambda_{s_0}\left(X_{T_0}X_{T_0}^t\right)- \varepsilon_{min} \label{lemme4:1}
		\end{align}
	\end{cor}

	\section{A greedy algorithm for column selection}
	
	The analysis in Section \ref{main} suggest that a greedy algorithm can be easily devised for efficient column extraction. The idea is quite simple: append the column which minimises the norm of the scalar products with the columns selected up to the current iteration. This algorithm is described with full details in Algorithm \ref{euclid} below. 
	
	\begin{algorithm}[ht]
		\caption{Greedy column selection}\label{euclid}
		\begin{algorithmic}[1]
			\Procedure{Greedy column selection}{}
			\State Set $s=1$ and choose a random singleton $T=\{j^{(1)}\} \subset \{1,\ldots,p\}$. Set $\eta^{(1)}=1$.
			\While{$\eta^{(s)} \ge 1-\epsilon$}  
			\State Set 
			\begin{align*}
			j^{(s)} & \in \textrm{argmin}_{j\in \{1,\ldots,p\}\setminus T} \quad \Vert X_T^t X_j\Vert_2.
			\end{align*}
			\State Set
			\begin{align*}
			\alpha^{(s)} & = \Vert X_T^tX_{j^{(s)}}\Vert_2^2/(s\mu^2).
			\end{align*}
			\State Set $T=T \cup \{j^{(s)}\}$.
			\State 
			\begin{align*}
			\eta^{(s+1)} & = \eta^{(s)}- \min \left({\sqrt{\alpha \ s}\mu, \frac{\alpha \mu^2 s}{1-\lambda_{s}(X_T^tX_T)}}\right)
			\end{align*}
			\State $s\gets s+1$
			\EndWhile\EndProcedure
		\end{algorithmic}
	\end{algorithm}
	Note that Algorithm 1 requires the computation of the smallest eigenvalue at each step, which might be computationally expensive in large dimensional settings.

	\section{Numerical experiments}
	\label{num}
	
	\subsection{Extracting representative time series}
\begin{figure}[H]
		\centering
		\includegraphics[scale=0.37]{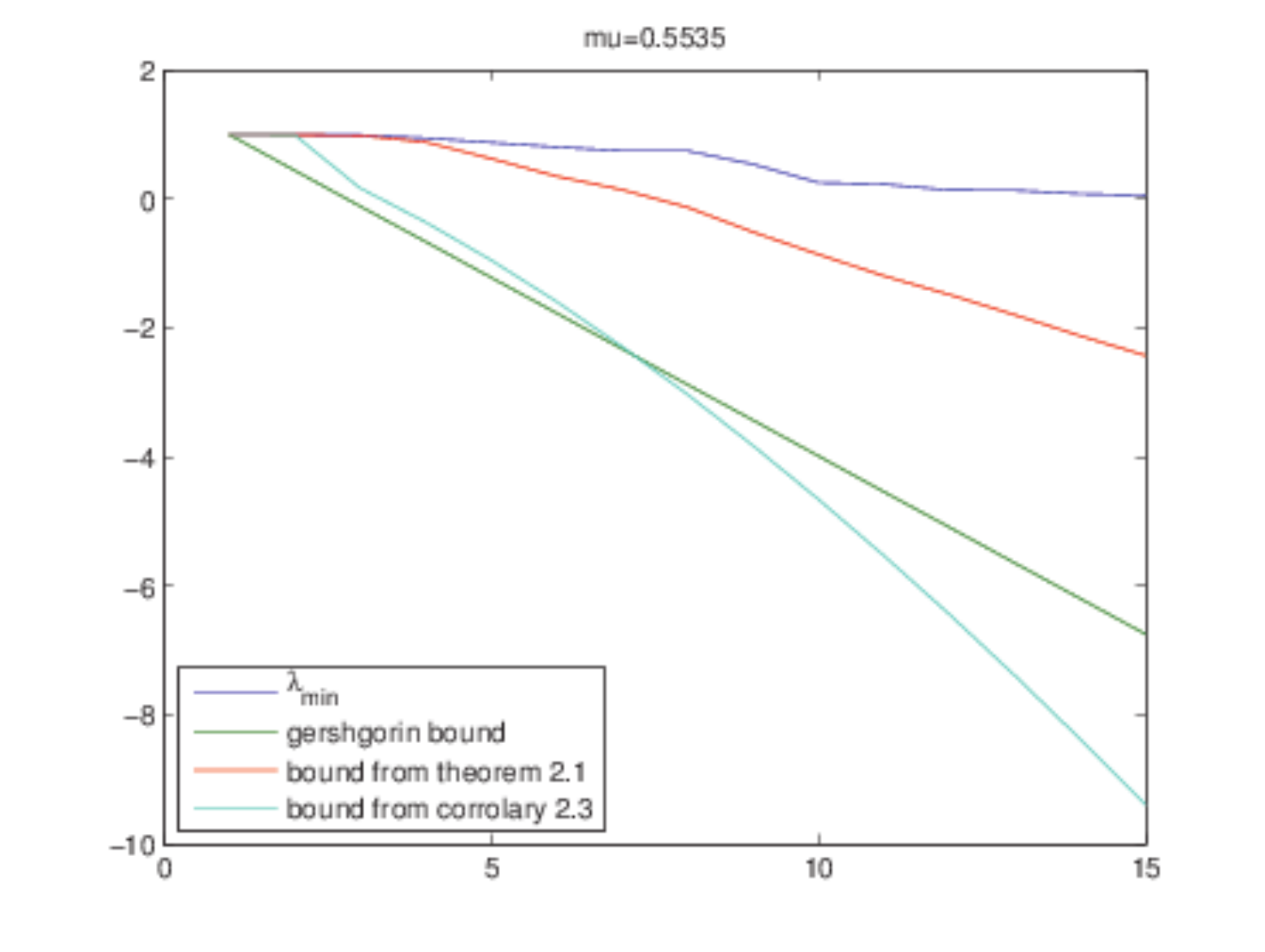}
		\includegraphics[scale=0.18]{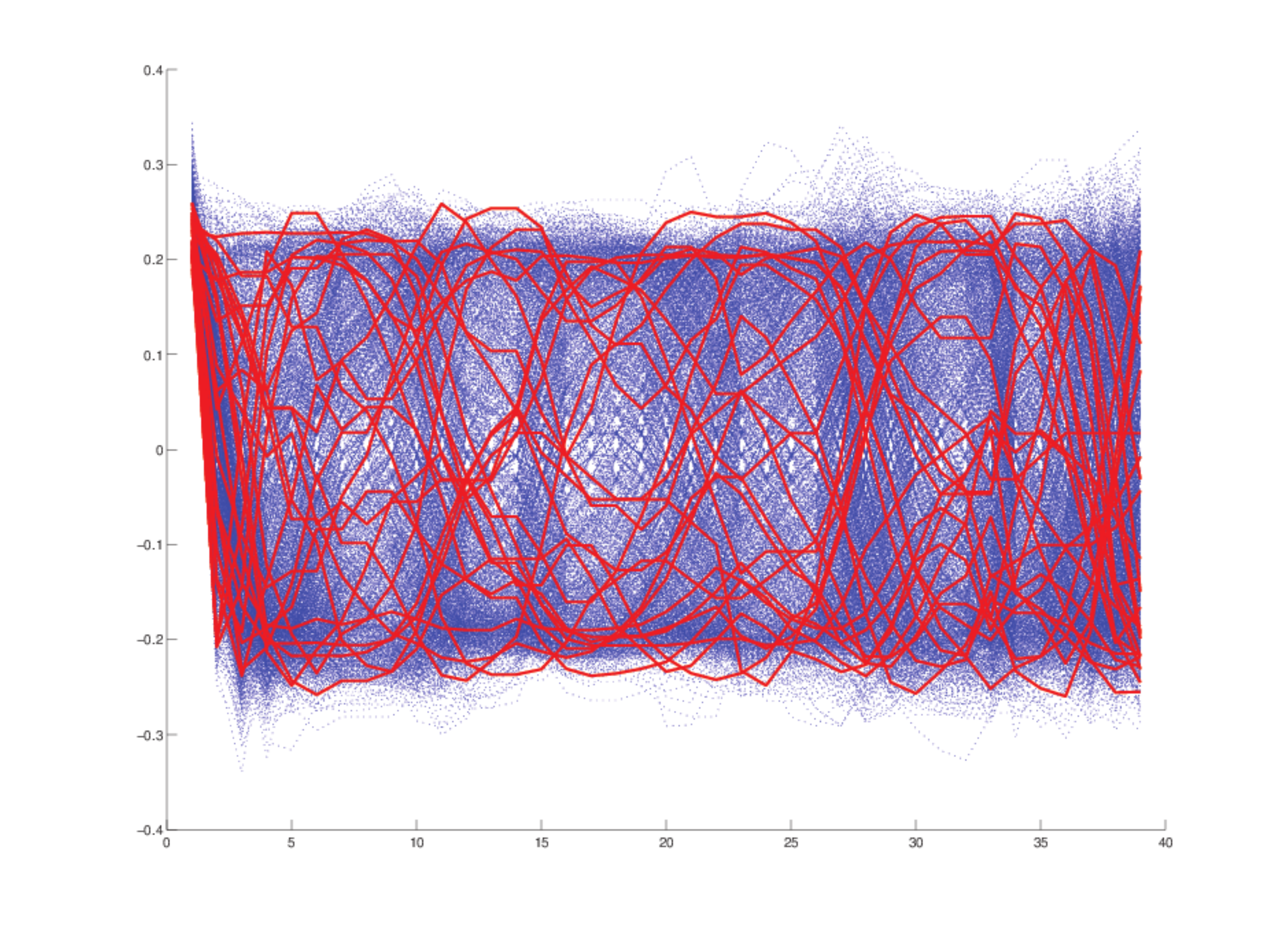}
		\caption{Left: Evolution of the smallest singular value in the greedy column selection Algorithm 1. Right: Main extracted Features.}
		\label{f1}
	\end{figure}

	Time series are ubiquitous in a world where so many phenomena are monitored via sensor networks. One interesting application of greedy column selection is to 
	\begin{itemize}
		\item extract representative time series among large datasets and 
		
		\vspace{.5cm}
		
		\item understand the intrinsic "dimension" of the dataset, i.e. the maximum number of different dynamics that are present.
		
		\vspace{.5cm}
		
		\item extract potential outliers.
	\end{itemize}
	In this experiment, we considered a set of 1479 times series of length 39 which consist in non-linear transformation of satellite InSAR data \footnote{a non-linear transformation was performed in order to make the time-series locations and sources impossible to identify}. 
	Then, starting from a random time series, we extracted 150 times series sequentially minimizing $\Vert X^t_T X_j\Vert_2, j\notin T$ at each step. Figure \ref{f1} shows the behavior of our algorithm over time. For large $\mu$, we see that the bound provided by Corollary \ref{cor3.4} are worse than the Gershgorin bound and successive applications of Theorem \ref{theorem:2} provides again a better bound. 
	
	\subsection{Extracting representative images from a dataset}
	
\begin{figure}[ht]
		\centering
		\includegraphics[width=10cm]{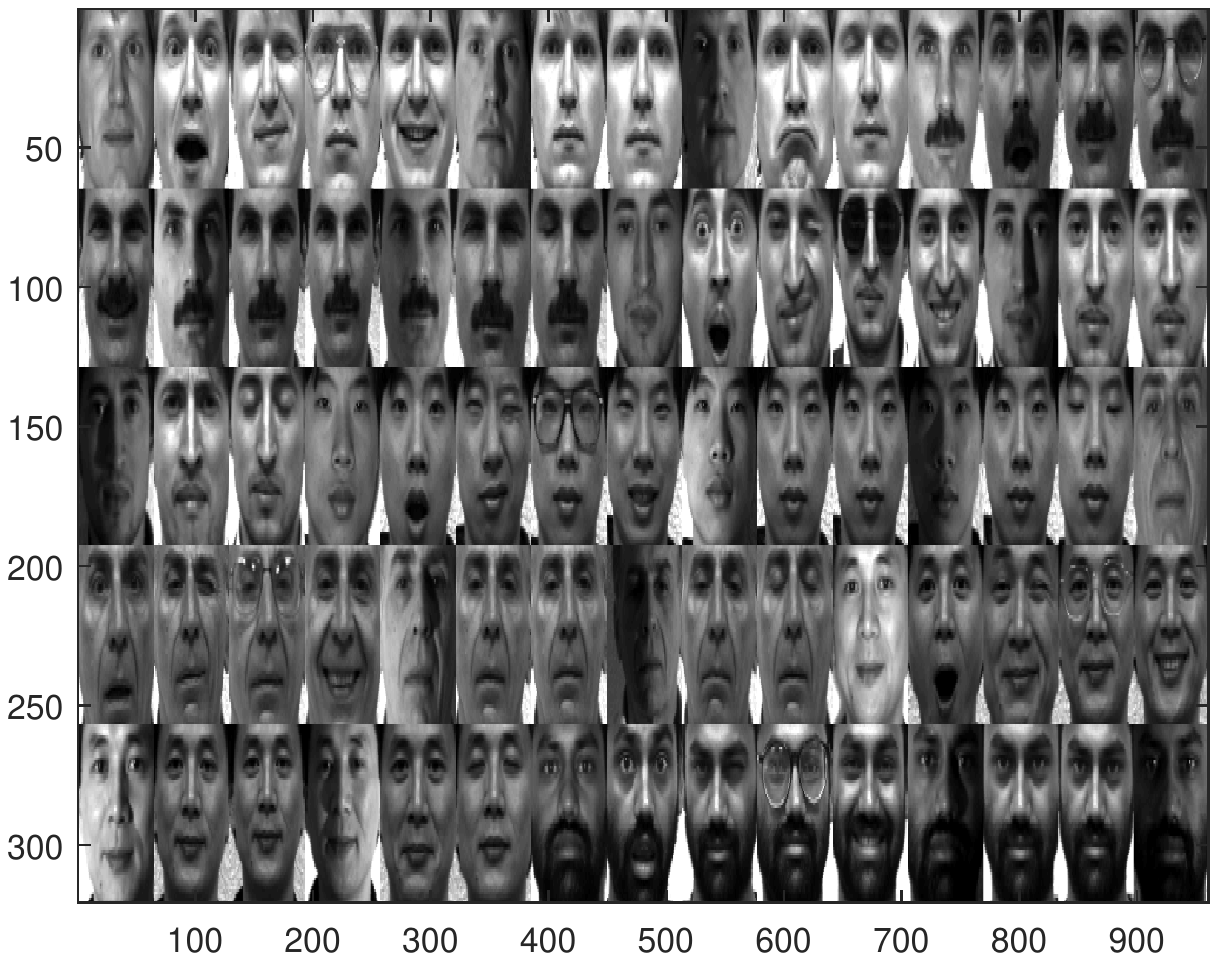}
		\includegraphics[width=6cm]{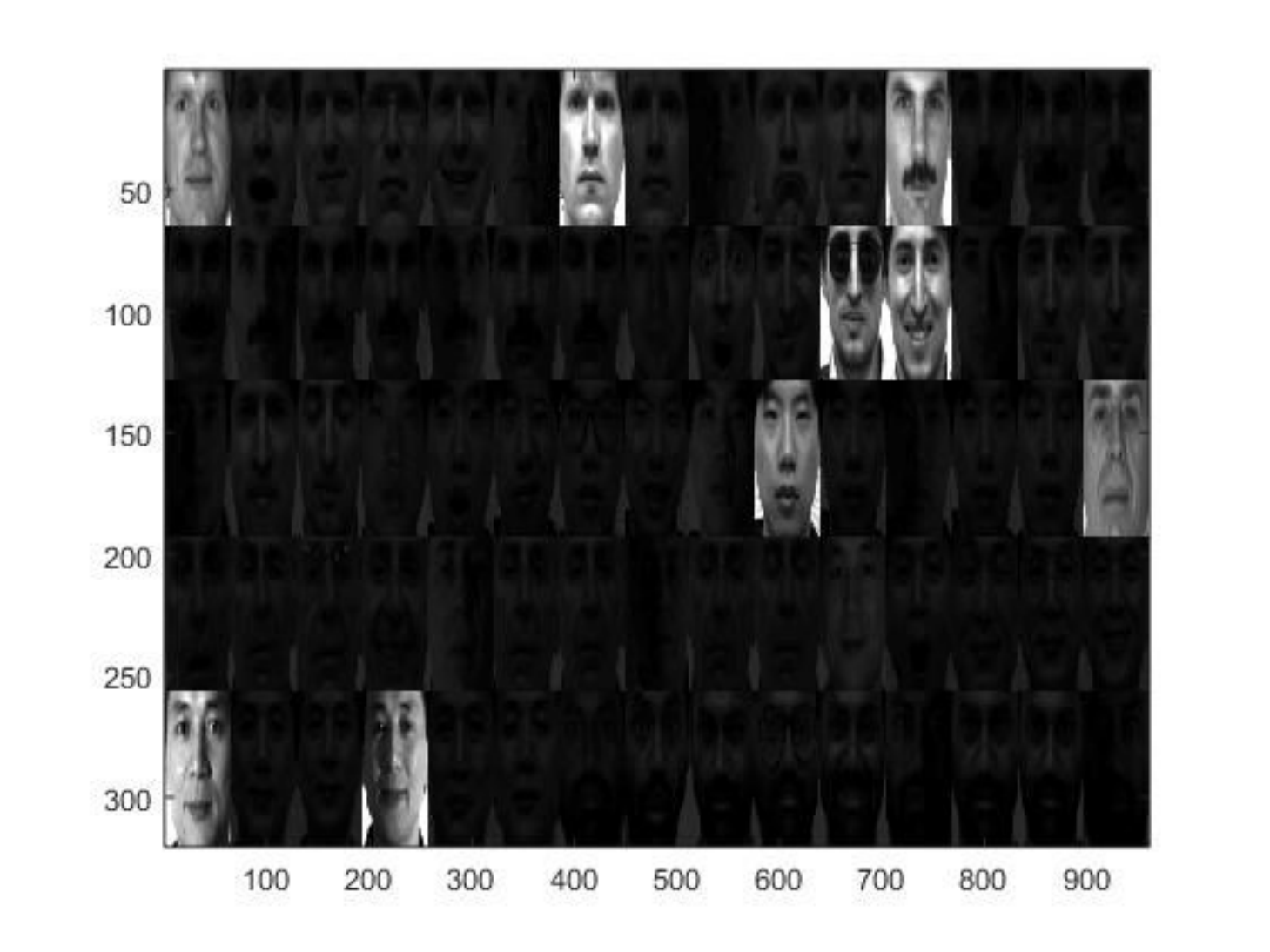}\caption{Left: Faces from the Yale database. Right: Faces selected by our algorithm.}
		\label{Yale}
	\end{figure}

	Extracting representative objects in a dataset is of great importance in data analytics. It can be used to detect outliers or clusters. In this example, we applied our technique to the Yale Faces database shown in Figure \ref{Yale} (Left). 
	In order to cluster the set of images, we performed a preliminary scattering transform \cite{mallat2012group}, \cite{bruna2013invariant} of the images in the dataset. We then reshaped the resulting scattering transform matrices into column vectors that we further  concatenated into a single matrix $X$.
	We selected 9 faces using our column selection algorithm and we obtained the result shown in Figure \ref{Yale} (Right). The total time for this computation was .07 seconds. Larger Pictures are given in the assiociated report \cite{chretienhoprep}.
	
	\subsection{Comparison with CUR}
	
\begin{figure}[ht]
		\centering
		\includegraphics[width=6cm]{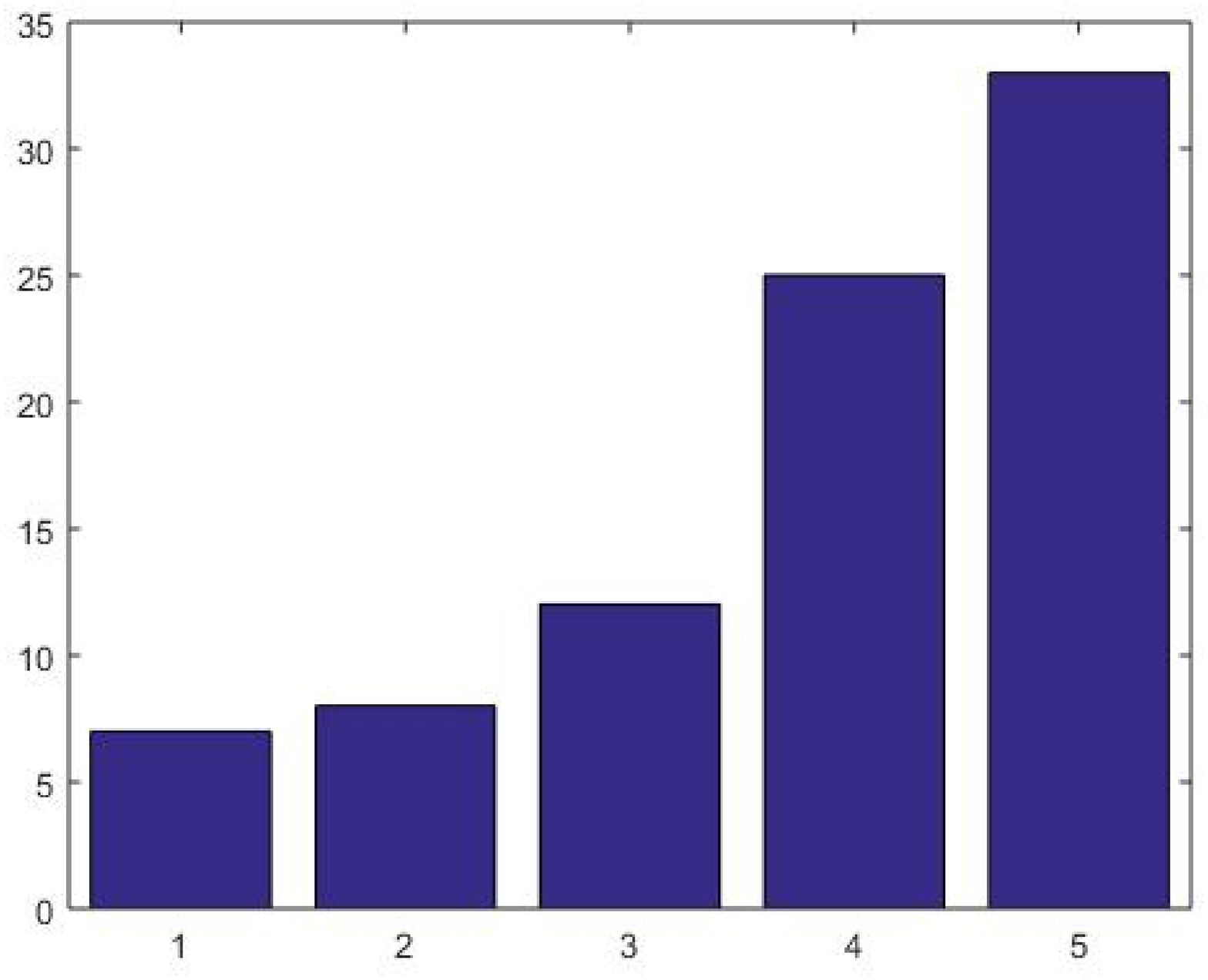}
		\includegraphics[width=6cm]{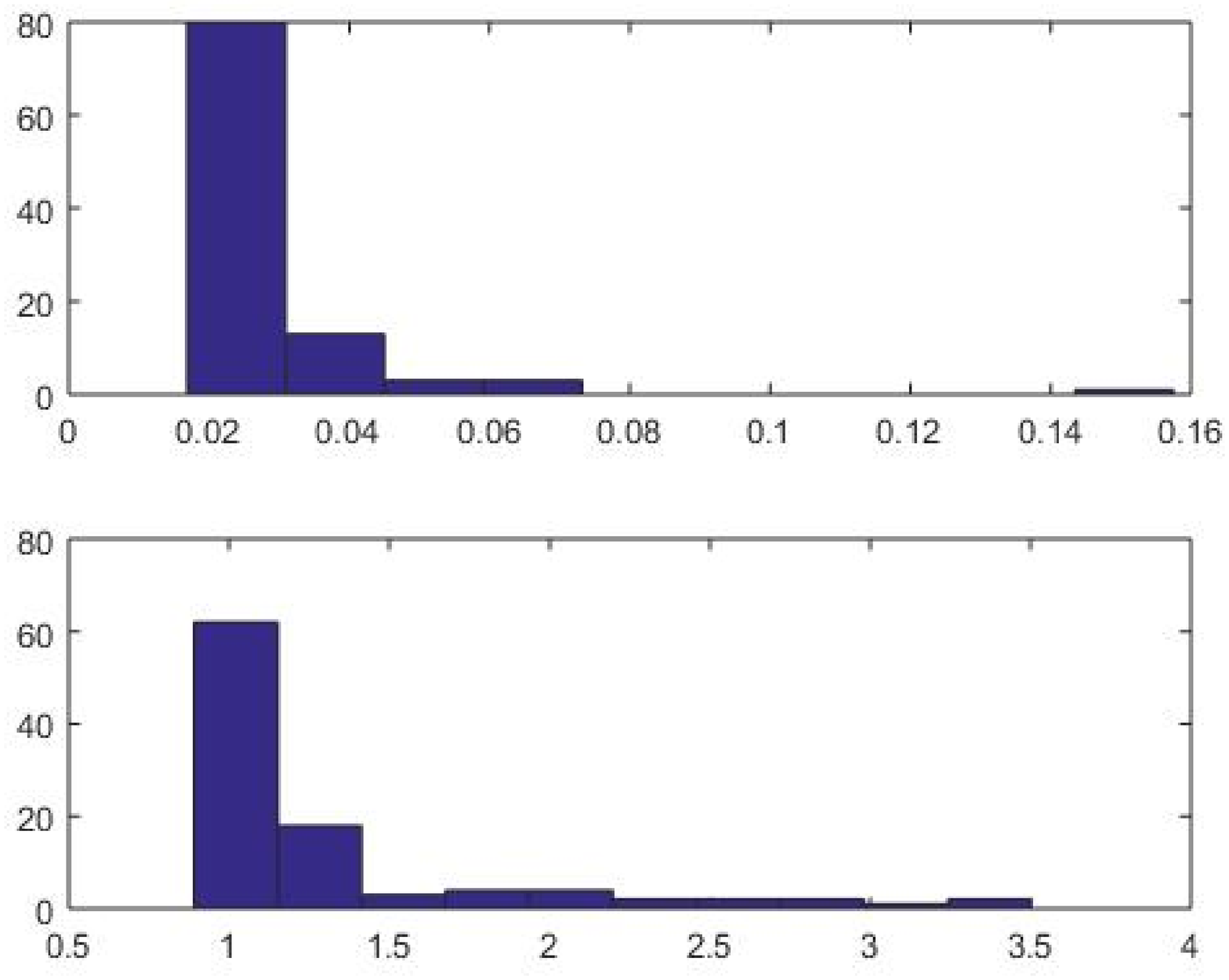}
		\caption{Left: counts of the number of singular values of the submatrix extracted with Algorithm 1 larger than for CUR among the 5 smallest singular spectrum for 100 independant Monte Carlo trials. Right-top: histogram of the computation time for Algorithm 1. Right-bottom: histogram of the computation time for the CUR method \cite{boutsidis2009improved}.}
		\label{fig:comps}
	\end{figure}

	We compared the behavior of our method with the CUR algorithm proposed in \cite{boutsidis2009improved}. We generated 100 matrices with i.i.d. standard Gaussian entries, with 100 rows and 10000 columns and performed both Algorithm 1 from the present paper and the CUR method. We restricted the study to the case of 10 columns to be extracted. The following histograms in Figure \ref{fig:comps} show the relative performance of our method as compared to CUR \cite{boutsidis2009improved} \footnote{we used the Matlab implementation provided on Christos Boutsidis webpage}. 
	
	The Monte Carlo experiments shown in Figure \ref{fig:comps} prove that our method preforms better than the CUR method, both from the viewpoint of providing submatrices with larger singular values on average and for a much smaller computational effort (our method was around 50 times faster for these experiments). These experiments are extracted from a more extensive set of experiments, including comparison with other methods, proposed in \cite{chretienhoprep}.

	\section{Conclusion and perspectives}
	In this paper, we established a relationship between the coherence and a perturbation bound for incoherent matrices. Our approach is based on perturbation theory and no randomness assumption on the design matrix is used to establish this property. 
	Coherence plays an important role in many pure and applied mathematical problems and perturbation results may help go significantly further. Two such problems for which we are planning further investigations are the following.
	
	\begin{itemize}
		\item \textbf{Random submatrices are well conditioned}.
		\label{pb1} Matrices with small coherence have a very nice property: most submatrices with $s$ columns have their eigenvalues concentrated around $1$ for $s$ of the order $n/\log(p)$. This was first studied in 	 \cite{tropp2008norms}, \cite[Theorem 3.2 and following comments]{candes2009near} and then improved in \cite{chretien2012invertibility}. The study of such properties is of tremendous importance in the study of designs for sparse recovery \cite{candes2009near}. An interesting potential application of studying spectrum perturbations after appending a column is the one of spectrum concentration via the bounded difference inequality \cite{boucheron2013concentration}. Such concentration bounds should also appear essential in understanding the behavior of random column sampling algorithms \cite{deshpande2010efficient}, \cite{boutsidis2014near}.
		
		\item \textbf{The restricted invertibility problem}.
		\label{pb2} Given any matrix $X$, the Restricted Invertibility problem of Bourgain and Tzafriri is the one of extracting the largest number of columns $X_j$, $j\in T$ form $X$ while ensuring that the smallest singular value of $X_T$ stays away from zero. Different procedures have been proposed for this problem. Some of them are randomised and some are deterministic. The original  results obtained by Bourgain and Tzafriri were based on random selection \cite{bourgain1987invertibility}. The current best results were recently obtained by Youssef in  \cite{youssef2014restricted} based on an remarkable inequality discovered by Batson, Spielman and Srivastava in \cite{batson2012twice}. In  \cite{chretien2015elementary}, using an elementary perturbation approach, the first author and S. Darses recently obtained a very short proof of a weaker version of the Bourgain-Tzafriri theorem (up to a $\log(s)$ multiplicative term). Our next goal is to  refine these types of perturbation results in the small coherence setting and extend the applicability to Big Data analytics.
	\end{itemize}

	\textit{Acknowledgements.}
	The work of the first author was funded by The National Measurement Office of the UK's Department for Business, Energy and Industrial Strategy supported this work as part of its Materials and Modelling programme.

	%\begin{biography}[{\includegraphics[width=1in,height=1.25in,clip,keepaspectratio]{mshell}}]{Michael Shell}
	% \begin{IEEEbiography}{St\'ephane Chr\'etien}
	% \end{IEEEbiography}

	%\newpage

	
	%\vfill
	
	
	\bibliographystyle{amsplain}
	\bibliography{biblio.bib}
	\appendix 
	
	\section{Interlacing and the characteristic polynomial}
	Recall that for a matrix 
	$A$ in $\mathbb R^{n\times n}$, $p_A$ denotes 
	the characteristic polynomial of $A$. 
	\begin{prop} {\bf Cauchy's Interlacing theorem}.
		If $A \in \mathbb R^{n\times n}$ is a symmetric matrix with 
		eigenvalues $\lambda_1 \geq \cdots \geq \lambda_n$ and associated eigenvectors $v_1$,\ldots,$v_n$, and $v \in \mathbb R^n$, then 	
		\begin{align}
		p_{A+vv^t}(x) & = p_A(x) \left(1-\sum_{i=1}^n \frac{\langle v,u_i \rangle^2}{x - \lambda_i} \right). 
		\end{align}
		\label{interlace}
	\end{prop} 
	The previous lemma states in particular that the eigenvalues of $A$ interlace those of $A+vv^t$.
	
	\section{Proof of Theorem \ref{cor3.4}}
	
	Define $\lambda_{s_0+s,\min}$ by
	\begin{align*}
	\left\{ \begin{array}{ll}
	\lambda_{s_0,\min}=\lambda_{s_0}\left(X_{T_0}X_{T_0}^t\right)& \\
	\lambda_{s_0+s+1,\min}=\lambda_{s_0+s}\left(X_{T_0\cup T}X_{T_0\cup T}^t\right)-\min\left(\sqrt{\alpha \mu^2 (s_0+s)}, \frac{\alpha \mu^2 (s_0+s)}{1-\lambda_{s_0+s,\min}}\right)&
	\end{array}
	\right.
	\end{align*}
	
	There are two step to prove for the theorem. The first step set up the basis for some recurrence relation. We show that, for $s\ge 0$, to obtain a lower-bound of $\lambda_{s_0+s+1}$, it is enough to use $\lambda_{s_0+s,\min}$ as the basis for Corrolarry \ref{cor3.3}. Or simply that we have
	
	\begin{align*}
	& \lambda_{s_0+s,\min}-\min\left(\sqrt{\alpha\mu^2(s_0+s)},\frac{\alpha(s_0+s)\mu^2}{1-\lambda_{s_0+s,\min}}\right) \\
	& \quad \le \lambda_{s_0+s}\left(X_{T_0\cup T}X_{T_0\cup T}^t\right)-\min \left(\sqrt{\alpha\mu^2 (s_0+s)},\frac{\alpha(s_0+s)\mu^2}{1-\lambda_{s_0+s}\left(X_{T_0\cup T}X_{T_0\cup T}^t\right)}\right) \\ & \quad \le \lambda_{s_0+s+1} \left(X_{T_{s+1}}X_{T_{s+1}}^t\right).
	\end{align*}
	
	It is obvious that the case where one minimum is equal to $\sqrt{\alpha \mu (s_0+s)}$ satisfy the property. Therefore, we study the following inequality
	
	\begin{align*}
	\lambda_{s_0+s,\min}-\frac{\alpha(s_0+s)\mu^2}{1-\lambda_{s_0+s,\min}} \le \lambda_{s_0+s}-\frac{\alpha(s_0+s)\mu^2}{1-\lambda_{s_0+s}\left(X_{T_0\cup T}X_{T_0\cup T}^t\right)}
	\end{align*}
	
	It is easily verified that the property is true for $s=0$. Denote
	
	\begin{align}
	\varepsilon& =\lambda_{s_0+s}\left(X_{T_0\cup T}X_{T_0\cup T}^t\right)-\lambda_{s_0+s+1}\left(X_{T_{s+1}}X_{T_{s+1}}^t\right).
	\end{align}
	
	Then the recursion step is equivalent to proving that
	
	\begin{align}\label{3.17}
	& \alpha \mu^2 \frac{s_0+s}{1-\lambda_{s_0+s}\left(X_{T_0\cup T}X_{T_0\cup T}^t\right)} +\alpha \mu^2 \frac{s_0+s+1}{1-\lambda_{s_0+s+1,\min}}\ge \varepsilon + \alpha
	\mu^2 \frac{s_0+s+1}{1-\lambda_{s_0+s}\left(X_{T_0}X_{T_0}^t\right)+\varepsilon}.
	\end{align}
	
	This inequality can be interpreted as the sum of errors obtained by applying Corollary \ref{cor3.3} twice is greater than the sum of errors obtained if we knew the true value after one perturbation then apply Corollary \ref{cor3.3}.
	
	Let $g$ be defined by
	
	\begin{align*}
	g_{s_0+s}(x)=x+\alpha \mu^2\frac{s_0+s+1}{1-\lambda_{s_0+s}(X_{T_0\cup T}X_{T_0\cup T}^t)+x}.
	\end{align*}
	
	Since $\varepsilon\le \alpha\mu^2\frac{s_0+s}{1-\lambda_{s_0+s}(X_{T_0\cup T}X_{T_0\cup T}^t)}$ by Corollary \eqref{cor3.3}, it is enough to prove $g$ increasing.
	
	A simple analysis show that $g$ is strictly increasing if \begin{align*}
	\alpha \mu^2 \frac{s_0+s+1}{(1-\lambda_{s_0+s}(X_{T_0\cup T}X_{T_0\cup T}^t))^2 }<\frac{3}{4}.
	\end{align*}
	
	In the case $\alpha \mu^2 \frac{s_0+s+1}{(1-\lambda_{s_0+s}(X_{T_0\cup T}X_{T_0\cup T}^t))^2}>\frac{3}{4}$, we can show that the left side of inequation \eqref{3.17} is larger than $1-\lambda_{s_0+s}(T_0\cup T)$ and this means that we obtain the trivial bound $0$ and therefore of not relevant interest.
	
	This solves the problem of not knowing the true value $\lambda_{s_0+s}(T_0\cup T_1)$.\\
	
	For the second part, we aim at bounding the sum of errors. We have
	
	\begin{align*}
	\sum\limits_{i=s_0}^{s_0+s}\min\left(\sqrt{\alpha \mu^2 i}, \frac{\alpha\mu^2 i}{1-\lambda_{i,\min}}\right)\le \min\left( \sum\limits_{i=s_0}^{s_0+s} \sqrt{\alpha \mu^2 i},\sum\limits_{i=s_0}^{s_0+s} \frac{\alpha\mu^2 i}{1-\lambda_{i,\min}}\right).
	\end{align*}
	
	The second sum writes
	
	\begin{align*}
	& \sum\limits_{i=s_0}^{s_0+s}\frac{\alpha \mu^2 i}{1-\lambda_{s_0}(X_{T_0}X_{T_0}^t)+\sum_{j=s_0}^{i-1}\frac{\alpha \mu^2 j}{1-\lambda_{s_0}(X_{T_0}X_{T_0}^t)}} \\ 
	& \quad \quad =\sum\limits_{i=s_0}^{s_0+s}\frac{\alpha\mu^2 i}{1-\lambda_{s_0}(X_{T_0}X_{T_0}^t)+\frac{\alpha \mu^2}{1-\lambda_{s_0}(X_{T_0}X_{T_0}^t)}\sum_{j=s_0}^{i-1}j}
	\end{align*}
	
	This is equal to
	
	\begin{align*}
	& \sum\limits_{i=s_0}^{s_0+s}\frac{\alpha \mu^2 i}{1-\lambda_{s_0}(X_{T_0}X_{T_0}^t)+\sum_{j=s_0}^{i-1}\frac{\alpha \mu^2 j}{1-\lambda_{s_0}(X_{T_0}X_{T_0}^t)}} \\ & \quad =\sum\limits_{i=s_0+1}^{s_0+s}\frac{\alpha\mu^2 i}{1-\lambda_{s_0}(X_{T_0}X_{T_0}^t)+\frac{\alpha \mu^2 s_0(i-1)}{1-\lambda_{s_0}(X_{T_0}X_{T_0}^t)}}+\frac{\alpha\mu^2 s_0}{1-\lambda_{s_0}(X_{T_0}X_{T_0}^t)}
	\end{align*}
	
	Simple computations lead to the result.
	
	Therefore applying $s_1$ times Corollary \ref{cor3.3} and each time upper-bounding, we have \eqref{lemme4:1}.

\end{document}